\documentclass{article}
\usepackage{amsmath,graphicx,mlspconf}

\title{PERSONALIZING SMARTWATCH BASED ACTIVITY RECOGNITION USING TRANSFER LEARNING}
\twoauthors
  {Karanpreet Singh\sthanks{Ph.D. Candidate, Kevin T. Crofton Department of Aerospace and Ocean Engineering, Virginia Tech, Blacksburg, VA 24060. The author performed the work while being Machine Learning Research Intern at Qeexo Co., Pittsburgh, PA 15206}}
	{Virginia Tech\\
	Blacksburg, VA 24060\\
  kasingh@vt.edu}
  {Rajen Bhatt, Ph.D.\sthanks{Director of Engineering, Machine Learning R\&D, Qeexo Co., Pittsburgh, PA 15206}}
	{Qeexo Co.\\
	Pittsburgh, PA 15206\\
  rajen.bhatt@qeexo.com}

\begin{document}
%

\maketitle
\begin{abstract}
Smartwatches are increasingly being used to recognize human daily life activities.  These devices may employ different kind of machine learning (ML) solutions. One of such ML models is Gradient Boosting Machine (GBM) which has shown an excellent performance in the literature.  The GBM can be trained on available data set before it is deployed on any device.  However, this data set may not represent every kind of human behavior in real life. For example, a ML model to detect elder and young persons running activity may give different results because of differences in their activity patterns.  This may result in decrease in the accuracy of activity recognition.  Therefore, a transfer learning based method is proposed in which user-specific performance can be improved significantly by doing on-device calibration of GBM by just tuning its parameters without retraining its estimators. Results show that this method can significantly improve the user-based accuracy for activity recognition.
\end{abstract}
\section{Introduction}
\label{sec:intro}

Over the last decade, human activity monitoring devices have seen an exceptional development. The motivation behind this is that monitoring human activity has become increasingly important because of multiple reasons. As \cite{jovanov2006wban} stated that there are many studies which link physical activity with overall health status. Moreover, recognizing activities such as walking, running, or cycling in patients with diabetes, obesity, heart disease, dementia or other mental pathologies helps in improvement in their treatments \cite{lara2013survey}.

There are different kinds of heath monitoring devices which have been studied in the literature such as pedometers, smartphones, smartwatches, etc. Chan \textit{et al.} \cite{chan2004health} studied the health benefits of a pedometer-based physical activity intervention in sedentary workers. They showed improvements in waist girth and resting heart rate in workers with increase in the number of accumulated steps per day. Smartwatches and smartphones are other options for human activity recognition. These devices contain accelerometers and gyroscopes that can help detecting one's activity being performed based on their body movements.

In literature, many Machine Learning (ML) solutions have been used for human activity recognition. Anguita \textit{et al.} \cite{anguita2012human} studied human activity recognition on smartphones using a multi-class Support Vector Machine (SVM). Baccouche \textit{et al.} \cite{baccouche2011sequential} presented a deep neural network to classify sequences of human actions. Ronao and Cho \cite{ronao2016human} presented a deep convolutional neural network (convnet) to perform efficient and effective human activity recognition using smartphone sensors.

Smartwatches have some advantages over smartphones in activity recognition. For example, Weiss\textit{ et al.} \cite{weiss2016smartwatch} stated that the smartphone could shift in a person’s pocket and pocket position is not ideal for tracking hand-based activities. They also stated that smartphone-based activity recognition has limitations for women because they generally do not keep the phones in their pockets. However, smartwatches help in addressing these issues and could perform better while making use of person's hand movements. 

Weiss\textit{ et al.} \cite{weiss2016smartwatch}  compared different ML algorithms using smartphones and smartwatches. They trained their ML models on two kinds of data: 1) data from only the intended user (called personal models) and 2) data from every user but the intended user (called impersonal models). They showed that personal models perform better than impersonal models but at the cost of requiring each user to provide labeled training data. Shahmohammadi \textit{et al. } \cite{shahmohammadi2017smartwatch} studied smartwatch based activity recognition using active learning. They stated that, in literature, many methods have been presented which even though achieve high accuracy on collected data set, but were not personalized for the user and required large data set. Therefore, they used active learning to develop user personalized models for activity recognition. These studies show that there is need to build personalized ML models for getting high accuracy for activity recognition.

In this paper, we propose a novel approach to personalize smartwatch based activity recognition for specific user. In this approach, we tune parameters of a trained Gradient Boosting Machine (GBM) without retraining its estimators. Initially, a GBM is trained on an available data set. Later, this GBM is used and its parameters are tuned on small data from a user to improve the activity recognition of that user's activities. Only a small activity data is required from the user to re-calibrate the GBM for improving the performance of GBM for that specific user. This technique is very beneficial for elders, young children and diseased patients as data collection could be difficult from these people for long time. The commercially available smartwatch tend to not to perform well on these people. Therefore, with this motivation, the authors present the current novel approach to resolve this issue.

\section{METHODOLOGY}
\label{sec:illust}

In this section, we present the methodology for personalizing the smartwatch based activity recognition using transfer learning. Initially, a general introduction of GBM is provided. The authors assume that the reader would have knowledge about methodology behind the training of estimators of GBM ML model as it is not covered in this article. For those details, \cite{friedman2001greedy} and \cite{natekin2013gradient} can be referred. Later, the algorithm for tuning GBM parameters for improving user-specific accuracy is presented.

\subsection{GRADIENT BOOSTING MACHINE (GBM)}

The conventional ensemble algorithms like Random Forests calculates the average of the outputs of models in the ensemble. However, the boosting ensemble methods are based on different constructive strategy. In boosting, we add new models to the ensemble sequentially during training. \cite{natekin2013gradient} After every iteration, a new weak learner model is trained with respect to the error of the whole ensemble learnt so far. 

Friedman \cite{friedman2001greedy} introduced Gradient Boosting Machine (GBM) algorithm in 2001. This algorithm sequentially fits new machine learning models to provide a better estimate of the response variable. \cite{natekin2013gradient} The main goal of this algorithm is to build weak learners to be maximally correlated with the negative gradient of the loss function related to the whole ensemble. 

\subsection{TUNING OF GBM PARAMETERS}

Once a GBM is trained, it is used to predict the output for any particular inputs. Initially, the decision functions from all the estimators is predicted. For example, the decision function, $\phi_{jp}$, is calculated from $j^{th}$ estimator for $p^{th}$ class. Later, a final score is calculated for $p^{th}$ class in a multi-class output as
\begin{equation}
f_{p} = \phi_{0p}+\sum_{j=1}^n w_{jp}\phi_{jp}
\end{equation}
where, $n$ is the total number of estimators, $\phi_{0p}$ is the initial guess for $p^{th}$ class, $\phi_{jp}$ is the decision output from $j^{th}$ estimator for $p^{th}$ class and $w_{jp}$ is the weight attached to this $j^{th}$ estimator. 

The score output for the each class is used to find the probability, $P(p)$, of $p^{th}$ class by using softmax function as

\begin{equation}
    P(p) = \frac{\exp (f_{p})}{\sum_{g=1}^{l} \exp(f_{g})}
\end{equation}

where, $l$ is the total number of classes. Finally, an output is predicted to be the class having the maximum probability.

The training of GBM could take could require significant computing resources and time. And, it is only possible to be trained on a limited available data set. For example, a GBM can be trained on data based on 'running' and 'walking' activities available from many users. However, this data set may not represent every kind of human behavior in real life. This may result in decrease in the accuracy of activity
recognition. Also, training a full personalized GBM on-device is also considered difficult as per current state of the art computing configurations. 

Therefore, we propose a transfer learning based approach where we use a GBM already trained on an available data set and later we just train the weights, $w_{jp}$, on small data set from a user to improve accuracy for that user specifically. During this approach, we do not alter the estimators of GBM. This approach has the benefit of being possible to be done on device. For example, a trained GBM can be deployed on a smartwatch and later, a new user would be asked to perform the required activities, like running and walking, while wearing the device. The new data collected from the user would be then used to train the weights, $w_{jp}$, of the GBM.

\begin{figure*}[t]
\begin{center}
\includegraphics[page=1,clip,trim={0.5cm 10cm 2cm 1cm},width=15cm]{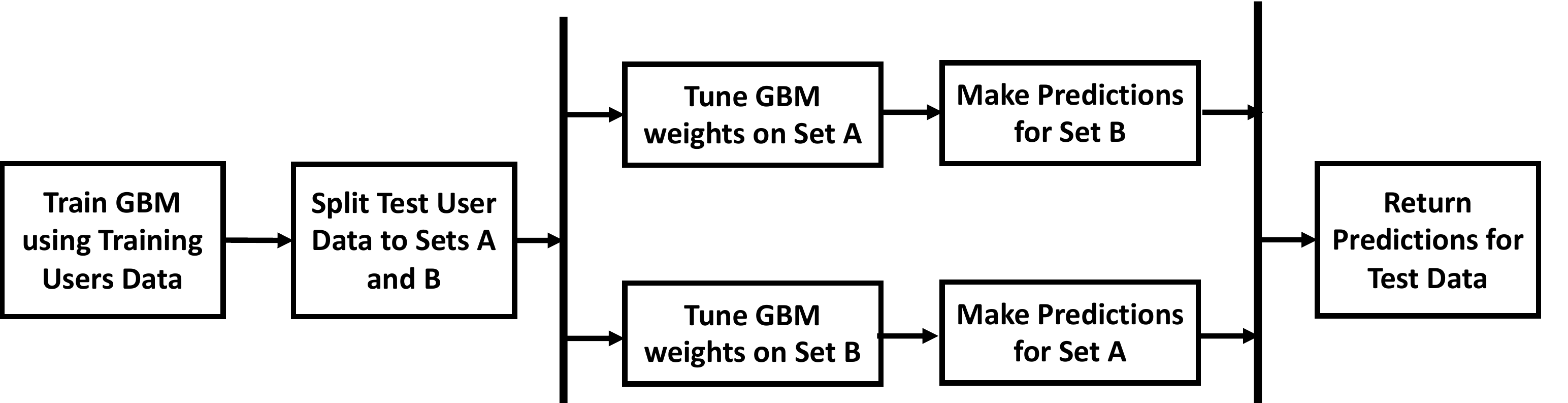}
\caption{Flowchart of Tuning of GBM Weights and Model evaluation}
\label{f:Flowchart}
\end{center}
\end{figure*}

To train the weights of the GBM, different loss functions can be used. For example, if mean squared error is used as the loss function to train the weights, the loss function can be defined as 

\begin{equation}
    L = \frac{1}{l} \sum_{p=1}^l (\hat{P}(p)-P(p))^2
\end{equation}

where, $\hat{P}(p)$ is the ground truth probability for $p^{th}$ class. This probability would be 1 for one of the classes and 0 for the remaining classes for any particular input. 

The gradient of the loss function with respect to weight $w_{jp}$ can be defined as

\begin{equation}
    \frac{\partial L}{\partial w_{jp}} =  \frac{2}{l} \sum_{p=1}^l (\hat{P}(p)-P(p)) (-\frac{\partial P(p)}{\partial w_{jp}})
\end{equation}

where, 

\begin{equation}
    \frac{\partial P(p)}{\partial w_{jp}} = P(p) \frac{\partial f_p}{\partial w_{jp}} - (P(p))^2 \frac{\partial f_p}{\partial w_{jp}}
\end{equation}

and

\begin{equation}
    \frac{\partial f_p}{\partial w_{jp}} = \phi_{jp}
\end{equation}
The gradient of the loss function can be used to update the weights as follows,

\begin{equation}
    \hat{w}_{jp} = w_{jp} - \eta \frac{\partial L}{\partial w_{jp}} 
\end{equation}

In the current work, stochastic gradient descent is used by making small batches of the small user data set to update the weights of the GBM. Therefore, batch size and learning rate are the hyper-parameters that can be tuned to improve accuracy on the validation set.

\section{APPLICATION AND RESULTS}

In this section, the methodology explained in Section 2 is applied on two publicly available data sets. The first data set is from \cite{altun2010comparative,barshan2014recognizing} and is called "Daily and Sports Activities Data Set". The second data set is from \cite{reiss2012introducing} and is called "PAMAP2 Data-set: Physical Activity Monitoring". 

In the current work, one-user-out cross-validation (CV) and F-1 scores are compared on these data sets for baseline and tuned GBM. Fig. \ref{f:Flowchart} shows the flowchart of the one-user-out CV procedure for tuning the GBM weights and model evaluation. The baseline GBM one-user-out CV is calculated by training GBM on $(N-1)$ users and tested on $N^{th}$ user. For tuning the GBM weights, the $N^{th}$ user's data is split to Sets A and B. Initially, GBM weights are tuned on Set A and then, the tuned GBM is used to make predictions for Set B and vice versa. This way, tuned GBM is used to calculate one-user-out CV. A part of the tuning data is used as a validation set to choose the final model based on validation set accuracy.

\subsection{Data Set I: Daily and Sports Activities Data-Set}

This data set has 19 different activities performed by 8 different subjects. The data is collected using accelerometer, gyroscopes, and magnetometers attached at different parts of body of the subjects. To show the potential of the presented approach in this paper, we selected four different activities: Running, Biking, Resting and Walking from this data set. Out of all the sensors, we only used the data from accelerometer sensor attached to arms/wrists. 

\subsubsection{Features Extraction} 

The data was collected at 25 Hz of sampling frequency. In the current work, 1 sec of latency is considered. This means that 25 samples, collected every second, are used to generate one instance of features.
Tab.~\ref{t:features} shows different features computed using these samples. These features are calculated using accelerometer data about $x$, $y$ and $z$ axes.

\begin{table}[h]
\caption{List of Features Used in Activity Classfication}
\begin{center}
\begin{tabular}{|c|}
\hline
\textbf{Features}  \\ \hline
Mean               \\ \hline
Standard Deviation \\ \hline
Skewness           \\ \hline
Auto-correlation   \\ \hline
Range              \\ \hline
Root Mean Square   \\ \hline
\end{tabular}
\end{center}
\label{t:features}
\end{table}

\subsubsection{Results}

\begin{figure}[h]
\includegraphics[page=3,clip,trim={0.3cm 0cm 19cm 9cm},width=9cm]{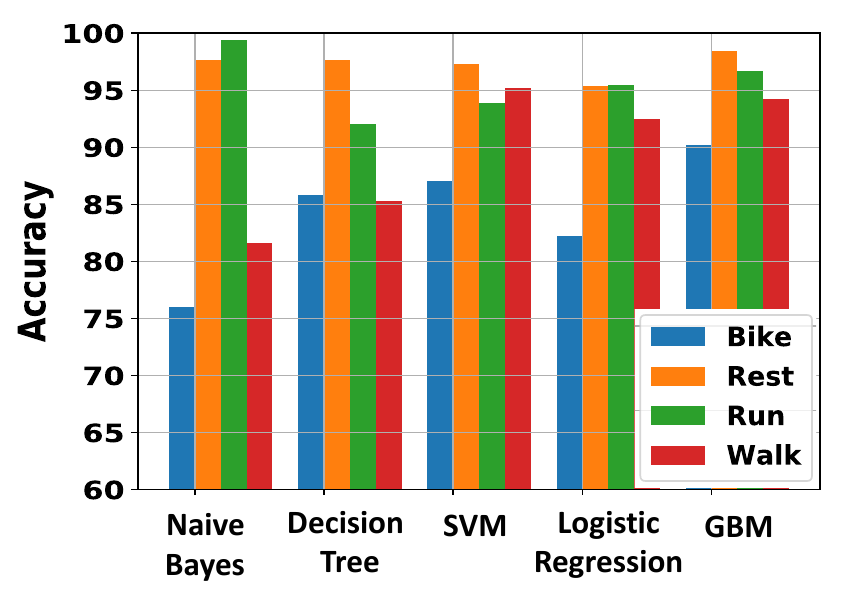}
\caption{Comparison of accuracy of GBM with other machine learning models on data set I}
\label{f:CompML}
\end{figure}

Initially, for comparison, the baseline GBM model one-user-out CV accuracy is compared with other ML models as shown in Fig.~\ref{f:CompML}. It is seen that for 'Rest' and 'Run' classes, every model has high accuracy. However, the GBM perform better than other models for 'Bike' and 'Walk' classes. This is also due to the reason that different users could walk and do cycling differently than other users.

\begin{figure*}[h]
\begin{center}
\includegraphics[page=2,clip,trim={0cm 5.5cm 0cm 0cm},width=16cm]{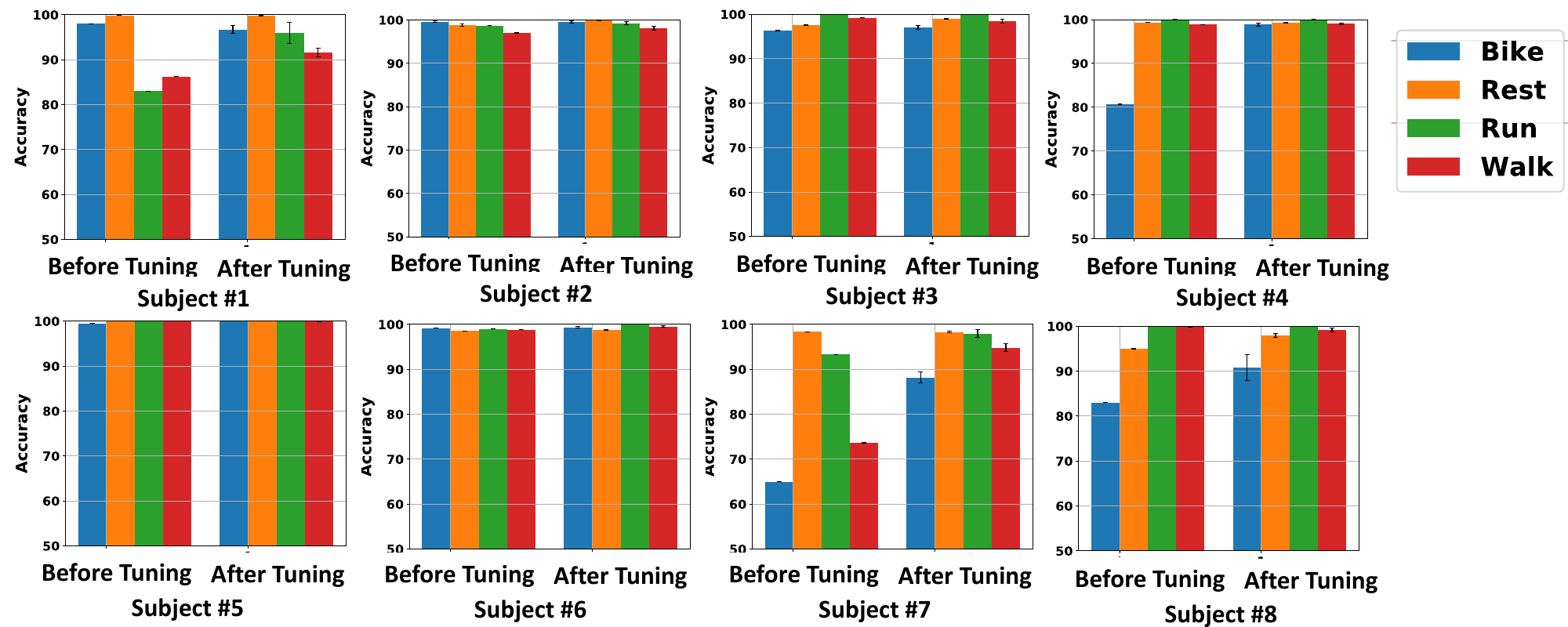}
\caption{Average increase in the accuracy of subjects in Data Set 1}
\label{f:SportsAccDetails2}
\end{center}
\end{figure*}

Later on, the tuning algorithm is used on this data set to improve one-user-out CV accuracy. Fig.~\ref{f:SportsAccDetails2} shows average increase in the accuracy of each subject in data set I. It can be seen that there is some increase in the accuracy of every subject. There is a significant increase in the accuracy of 'Bike' and 'Walk' class for subject \#7 from 65.00\% and 73.65\% to 88.15\% and 94.80\% respectively. Also, for subject \#8, there is an increase of accuracy for 'Bike' class from 83.00\% to 90.79\%.

\begin{figure}[h]
\includegraphics[page=3,clip,trim={15cm 0.5cm 5cm 9cm},width=9cm]{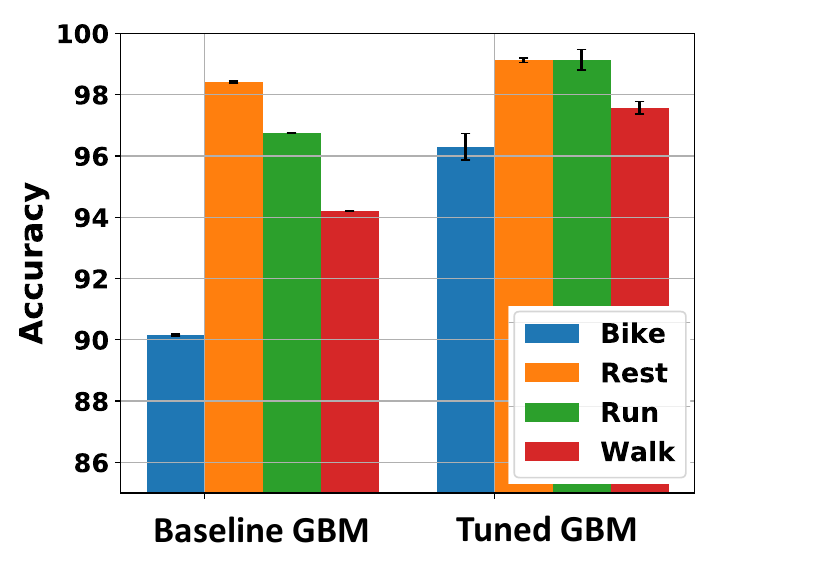}
\caption{Overall accuracy of tuned GBM on data set 1}
\label{f:OverallACC1}
\end{figure}

Fig.~\ref{f:OverallACC1} shows average increase in the overall one-user-out CV accuracy for each class after tuning the baseline GBM. The baseline accuracy is 90.14\%, 98.41\%, 96.75\% and 94.20\% for 'Bike', 'Rest', 'Run' and 'Walk' classes respectively. These accuracy increase to 96.30\%, 99.12\%, 99.14\% and 97.57\%. This shows that there is more than 50\% error reduction is possible by tuning the GBM on specific user's data. 

\begin{figure*}[h]
\begin{center}
\includegraphics[page=3,clip,trim={0cm 10cm 0cm 0cm},width=16cm]{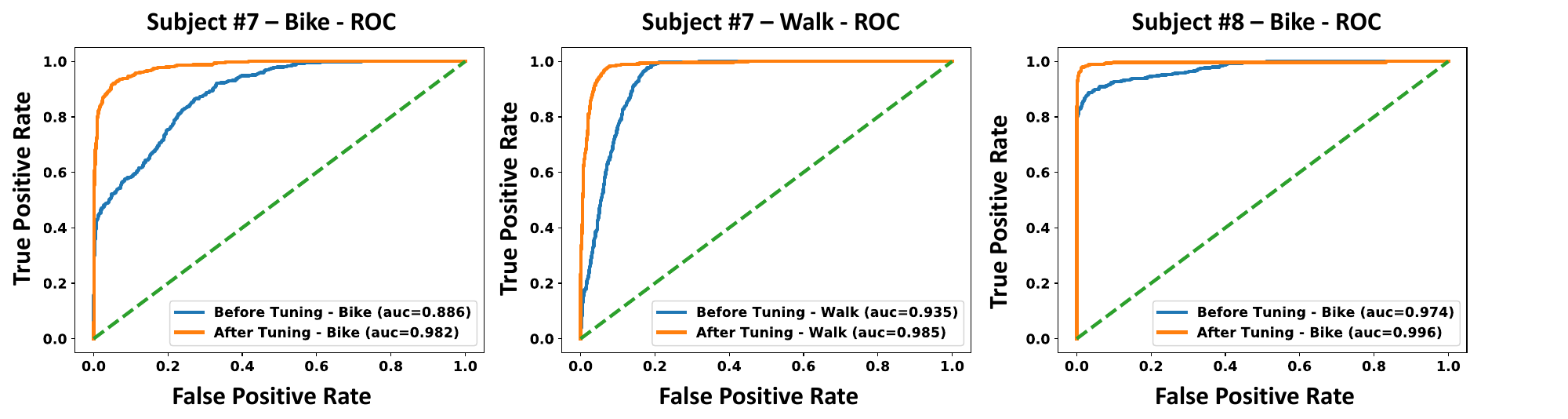}
\caption{Comparison of ROC curves of subject \#7 and \#8 from data set I for baseline and tuned GBM }
\label{f:ROCCUrves}
\end{center}
\end{figure*}

The ROC curves for subject \#7 and \#8 for 'Bike' and 'Walk' classes are shown in Fig.~\ref{f:ROCCUrves}. It can be seen that Area Under Curve (AUC) increases significantly from 0.886 to 0.982 for subject \#7 'Bike' class. Tab.~\ref{f:SportsF1} shows the comparison of F-1 scores of baseline GBM and tuned GBM. The overall F-1 score increase from 0.9456 to 0.9758. 

\begin{table}[h]
\begin{center}
\caption{Comparison of F1 scores of baseline and tuned GBM across subjects of data set I}
\begin{tabular}{|l|c|c|c|c|}
\hline
\textbf{Subject} & \textbf{Baseline} & \textbf{} & \textbf{Tuned GBM} & \textbf{\begin{tabular}[c]{@{}c@{}}Standard\\ Deviation\end{tabular}} \\ \hline
\#1              & 0.9231            &           & 0.9549             & 0.0044                                                                \\ \hline
\#2              & 0.9834            &           & 0.9904             & 0.0015                                                                \\ \hline
\#3              & 0.9817            &           & 0.9842             & 0.0019                                                                \\ \hline
\#4              & 0.9453            &           & 0.9917             & 0.0010                                                                \\ \hline
\#5              & 0.9986            &           & 0.9997             & 0.0002                                                                \\ \hline
\#6              & 0.9883            &           & 0.9935             & 0.0008                                                                \\ \hline
\#7              & 0.8004            &           & 0.9440             & 0.0042                                                                \\ \hline
\#8              & 0.9431            &           & 0.9685             & 0.0066                                                                \\ \hline
Overall          & 0.9456            &           & 0.9784             & 0.0011                                                                \\ \hline
\end{tabular}
\label{f:SportsF1}
\end{center}
\end{table}

\begin{figure*}[h]
\begin{center}
\includegraphics[page=4,clip,trim={0cm 5.3cm 0cm 0cm},width=16cm]{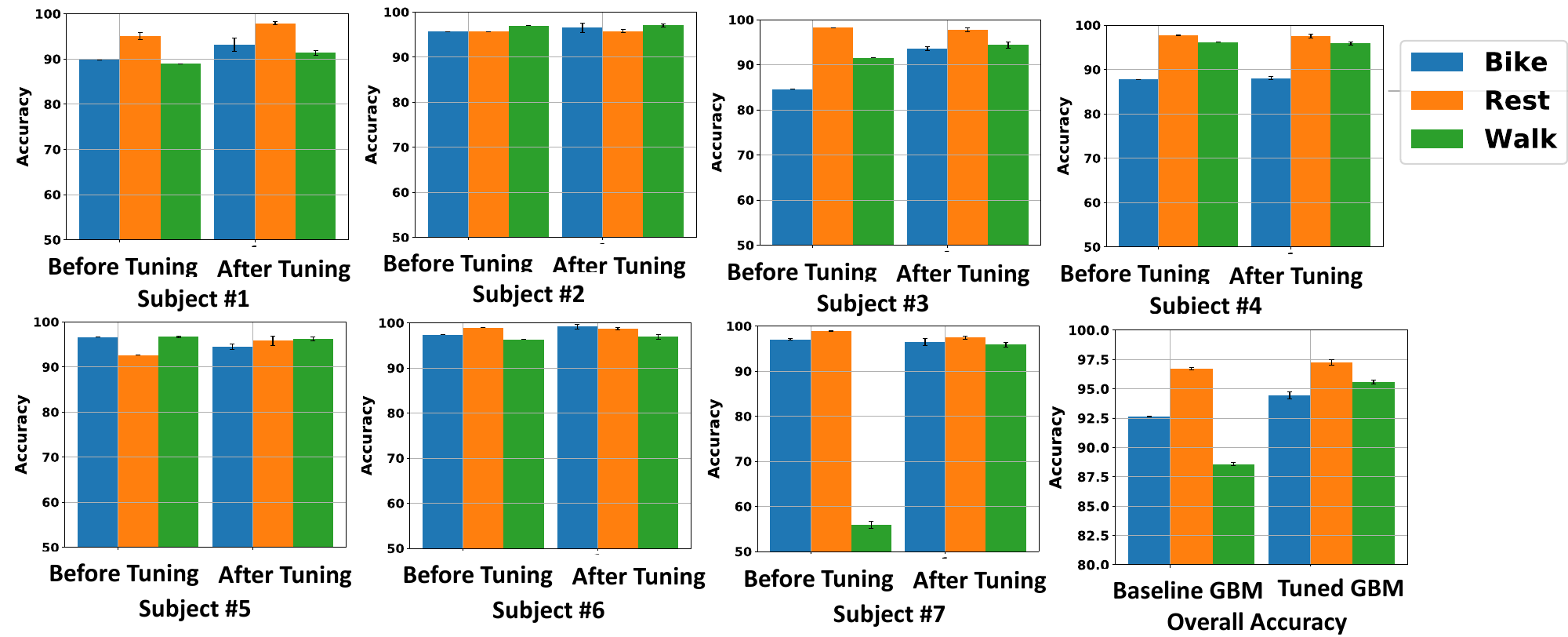}
\caption{Average increase in the accuracy of subjects in data set II}
\label{f:PAMAPAccDetails}
\end{center}
\end{figure*}

\subsection{Data Set II: PAMAP2 Data-set}

This data set has 12 different activities performed by 9 different subjects. Out of many different sensors used, we extracted only accelerometer data attached at the subject's wrist. Three different activities: Biking, Resting and Walking are considered in this study. We only used 7 subjects' data due to missing data of 2 subjects for the activities of our interest.

\subsubsection{Features Extraction}

This data set was collected at 100 Hz. As mentioned in above section, we again considered 1 sec of latency, meaning, we used 100 samples are used to calculate features instances. Similar features are calculated as mentioned in Section 3.1.1.

\subsubsection{Results}

In this section, the baseline GBM trained on data set II is used to tune its weights for improving the performance. Again, the flowchart in Fig.~\ref{f:Flowchart} is used to find on-user-out CV accuracy. Fig.~\ref{f:PAMAPAccDetails} shows average increase in the one-user-out CV accuracy of each class of subjects and overall increase in the CV accuracy by tuning the GBM.

Tab.~\ref{t:PAMAPF1} shows the comparison of F-1 scores for baseline and tuned GBM. The overall F-1 score increase from 0.9307 to 0.9619. There is a significant increase in the subject F-1 score of \#7 from 0.8117 to 0.9671. Fig.~\ref{f:PAMAPAccDetails} shows that 'Walk' accuracy of subject \#7 increase from 55.93\% to 95.90\%. The figure also shows the overall one-user-out CV increase for each class. The baseline accuracy are 92.65\%, 96.69\% and 88.57\% for 'Bike', 'Rest' and 'Walk' classes respectively. They increase to 94.45\%, 97.25\% and 95.56\%, thus, showing the potential of the current approach.

\begin{table}[h]
\begin{center}
\caption{Comparison of F1 scores of baseline and tuned GBM across subjects of data set II}
\begin{tabular}{|l|c|c|c|c|}
\hline
\textbf{Subject} & \textbf{Baseline} & \textbf{} & \textbf{Tuned GBM} & \textbf{\begin{tabular}[c]{@{}c@{}}Standard\\ Deviation\end{tabular}} \\ \hline
\#1              & 0.9231            &           & 0.9508             & 0.0025                                                                \\ \hline
\#2              & 0.9612            &           & 0.9634             & 0.0029                                                                \\ \hline
\#3              & 0.9368            &           & 0.9588             & 0.0020                                                                \\ \hline
\#4              & 0.9551            &           & 0.9535             & 0.0017                                                                \\ \hline
\#5              & 0.9468            &           & 0.9576             & 0.0047                                                                \\ \hline
\#6              & 0.9759            &           & 0.9807             & 0.0020                                                                \\ \hline
\#7              & 0.8117           &           & 0.9671             & 0.0021                                                                \\ \hline
Overall          & 0.9307            &           & 0.9619             & 0.0009                                                                \\ \hline
\end{tabular}
\label{t:PAMAPF1}
\end{center}
\end{table}

\section{CONCLUSION}

In this paper, we propose a novel approach to personalize smartwatch based activity recognition for a specific user. The algorithm only tune the weights of a trained GBM without retraining its estimators. The approach has the benefit of being possible to be done on device. Only a small amount of activity data is required from the user to re-calibrate the GBM for improving the performance of GBM for that specific user. The approach has shown significant increase in the accuracy and F-1 scores for the users who have low accuracy for certain activity classes. 

In future, the approach can be applied on other activities or gesture recognition for creating user personalized machine learning models.



\bibliographystyle{IEEEbib}
\bibliography{strings}

\end{document}